\documentclass[runningheads]{llncs}
\usepackage[T1]{fontenc}
\usepackage{graphicx}
\usepackage{authblk}
\usepackage{hyperref}
\usepackage{booktabs}
\usepackage{xcolor}
\usepackage{graphicx,subfigure,psfrag} 
\begin{document}
\title{SafeGen: Embedding Ethical Safeguards in Text-to-Image Generation}
%
%
\author{Dang Phuong Nam \inst{1, 2}\orcidID{0009-0008-0233-9979} \and
Nguyen Kieu Linh\inst{2}\thanks{Corresponding author: linhnk@ptit.edu.vn}\orcidID{0009-0005-6854-1350} \and Pham Thanh Hieu\inst{2}\orcidID{0009-0002-2786-2834}}
\authorrunning{D. P. Nam, N. K. Linh, and P. T. Hieu}
%
\institute{Bank for Investment and Development of Vietnam (BIDV), Hanoi, Vietnam\and Posts and Telecommunications Institute of Technology, Hanoi, Vietnam
\email{phuongnamdpn2k2@gmail.com,}
\email{linhnk@ptit.edu.vn,}
\email{hieupt@ptit.edu.vn}}
\maketitle              
\begin{abstract}
Generative Artificial Intelligence (AI) has created unprecedented opportunities for creative expression, education, and research. Text-to-image systems such as DALL·E, Stable Diffusion, and Midjourney can now convert ideas into visuals within seconds, but they also present a dual-use dilemma, raising critical ethical concerns: amplifying societal biases, producing high-fidelity disinformation, and violating intellectual property. This paper introduces SafeGen, a framework that embeds ethical safeguards directly into the text-to-image generation pipeline, grounding its design in established principles for Trustworthy AI. SafeGen integrates two complementary components: BGE-M3, a fine-tuned text classifier that filters harmful or misleading prompts, and Hyper-SD, an optimized diffusion model that produces highfidelity, semantically aligned images. Built on a curated multilingual (English-Vietnamese) dataset and a fairness-aware training process, SafeGen demonstrates that creative freedom and ethical responsibility can be reconciled within a single workflow. Quantitative evaluations confirm its effectiveness, with Hyper-SD achieving IS = 3.52, FID = 22.08, and SSIM = 0.79, while BGE-M3 reaches an F1-Score of 0.81. An ablation study further validates the importance of domain-specific fine-tuning for both modules. Case studies illustrate SafeGen’s practical impact in blocking unsafe prompts, generating inclusive teaching materials, and reinforcing academic integrity.

\end{abstract}
\section{Introduction}

The rapid proliferation of generative Artificial Intelligence (AI) has transformed the landscape of creative production, education, and scientific research. Among the most visible advances are text-to-image systems such as DALL·E, Stable Diffusion, and Midjourney \cite{ref1,ref2}. These models can translate abstract concepts into vivid imagery within seconds, offering new opportunities for teaching and learning. For educators, they provide a means to visualize difficult theories, design tailored illustrations, and stimulate student engagement. For researchers, they open avenues for creative experimentation, data visualization, and cross-disciplinary collaboration.

Yet these possibilities are inseparable from a set of profound ethical concerns. Because text-to-image models are trained on massive, uncurated datasets harvested from the Internet, they frequently inherit and amplify existing societal biases, producing images that reinforce stereotypes of gender, race, or cultural identity \cite{ref4,ref6}. At the same time, their ability to generate photorealistic but fabricated content raises the specter of large-scale disinformation and deepfake production, undermining trust in digital media \cite{ref10}. Questions of intellectual property add another layer of complexity: models may inadvertently reproduce copyrighted elements without attribution or permission \cite{ref11}. These risks are particularly acute in academic settings, where originality, accountability, and integrity are {foundational values} \cite{ref13}.

Attempts to address these problems through post-hoc safety mechanisms have proven inadequate. Commercial systems must walk a fine line between protecting users and preserving creative freedom. In practice, most solutions fall short. Stable Diffusion’s initial safety filter, for example, was criticized for its narrow scope: it targeted sexually explicit material but ignored violent or disturbing imagery \cite{ref14}. Conversely, Midjourney’s stricter filters were soon shown to be vulnerable to adversarial prompt attacks, in which slight modifications to input text successfully bypassed restrictions \cite{ref22}. {Current systems thus rely heavily on after-the-fact filtering, a brittle approach that is easily circumvented. What is needed instead is a framework that embeds ethical safeguards into the design of the generative process itself.}

This call for proactive solutions aligns with a broader international conversation on the governance of AI. Over the past decade, scholars and policy bodies—including the European Union, the OECD, and Floridi and Cowls—have articulated converging principles for Trustworthy AI \cite{ref19,ref20,ref21}. Across these initiatives, several themes recur: the imperative of fairness and inclusion, the obligation to prevent harm, the need for transparency and interpretability, the demand for accountability and human oversight, and the requirement of robustness in both technical and social contexts. These principles provide not only a normative compass but also a practical foundation for guiding the deployment of generative AI in sensitive environments such as education and research.

Generative AI directly tests the resilience of these principles. Bias amplification challenges fairness; deepfake production threatens non-maleficence; opaque safety filters undermine transparency; and the lack of clear governance weakens accountability. Academic contexts further demand attention to the integrity of scholarship, since the misuse of generative models risks eroding standards of originality and responsible authorship. Bridging the gap between principle and practice therefore requires concrete technical frameworks that operationalize ethical safeguards at the core of generative pipelines.

In response to this need, we introduce SafeGen, a framework that integrates ethical safeguards directly into text-to-image generation. SafeGen is grounded in the principles of Trustworthy AI and is designed specifically for educational and research contexts. The framework incorporates two core components: (i) BGE-M3, a fine-tuned classifier that proactively screens prompts for ethically problematic content before image synthesis; and (ii) Hyper-SD, an optimized diffusion model that produces high-quality, semantically aligned images while respecting ethical constraints. Together, these components demonstrate that creative freedom and ethical responsibility need not be mutually exclusive, but can be reconciled within a unified workflow.

{
	In parallel, other research has explored integrating safety mechanisms directly into text-to-image models. 
	For instance, Li et al.’s system (also called SafeGen) fine-tunes a diffusion model’s internal layers to eliminate sexually explicit visual concepts \cite{ref46}. 
	Similarly, Schramowski et al.’s Safe Latent Diffusion method removes or suppresses inappropriate image content during the generation process without requiring additional training \cite{ref47}. 
	However, such methods typically focus on a narrow subset of unsafe content (primarily NSFW material) or address only one aspect of the problem. 
	By contrast, our SafeGen targets a broader spectrum of ethical risks (from bias and hate to misinformation and academic misconduct) and employs a dual-module strategy, combining prompt filtering with bias-aware generation, to provide a more comprehensive solution tailored to educational and research environments.}

\section{Establishing the five pillars of SafeGen} \label{section2}
SafeGen is grounded in widely recognized frameworks for Trustworthy AI, including the EU’s Ethics Guidelines for Trustworthy AI \cite{ref20}, OECD AI Principles \cite{ref21}, UNESCO’s Recommendation on the Ethics of AI \cite{ref42}, and academic works such as Floridi and Cowls’ unified framework \cite{ref19}. From these foundations, we adapt five actionable pillars tailored to the academic use of generative AI:
\begin{itemize}
	\item[1.] \textit{Fairness, Non-Discrimination, and Inclusion:}
	Rooted in the EU’s fairness principle \cite{ref20} and OECD’s human rights emphasis \cite{ref21}, this pillar targets algorithmic bias. SafeGen aims to ensure equitable outcomes for all users, reducing stereotypes and reinforcing inclusion \cite{ref19}.
	
	\item[2.] \textit{Prevention of Harm and Promotion of Well-Being:}
	Combining the EU’s safety requirements \cite{ref20}, OECD’s focus on societal well-being \cite{ref21}, and UNESCO’s “do no harm” principle \cite{ref42}, this pillar emphasizes preventing harmful content—including misinformation, violence, or hate speech—while promoting constructive educational use.
	
	\item[3.] \textit{Transparency and Interpretability:}
	A common theme across global frameworks \cite{ref19,ref20,ref21}, this pillar highlights the need for systems to remain intelligible to users. SafeGen not only provides clear explanations when prompts are blocked but also communicates the model’s capabilities and limitations. This connects the framework to the broader agenda of \textit{Explainable and Interpretable AI}, which emphasizes human-understandable justifications for AI decisions and promotes user trust \cite{ref41,ref42}.
	
	\item[4.] \textit{Accountability and Human Oversight:}
	Following EU and OECD requirements for accountability and oversight \cite{ref20,ref21}, this pillar establishes governance mechanisms, documentation standards, and human-in-the-loop processes. It ensures that AI decisions can be audited, challenged, and corrected when necessary \cite{ref42}.
	
	\item[5.] \textit{Robustness, Security, and Academic Integrity:}
	Inspired by the robustness principle of EU and OECD \cite{ref20,ref21}, this pillar extends beyond technical resilience against adversarial attacks. It also encompasses social robustness, ensuring that SafeGen supports the preservation of academic standards such as originality, integrity, and responsible authorship \cite{ref44}.
\end{itemize}

Together, these five pillars provide a normative foundation that aligns SafeGen with international principles for Trustworthy AI while adapting them to the specific challenges of generative models in academic environments.

\section{The SafeGen framework: System design and data foundations}

SafeGen is designed as a practical instantiation of the five ethical pillars established in Section \ref{section2}. Its architecture ensures that fairness, harm prevention, transparency, accountability, and robustness are embedded not as optional add-ons but as integral design choices throughout the system.

\subsection{Conceptual Design}

The core principle of SafeGen is that ethical safeguards must precede and accompany the generative process. To this end, the framework integrates two complementary modules:

\begin{itemize}
	\item \textit{BGE-M3 Classifier:} A fine-tuned Transformer-based model that proactively screens prompts, filtering out harmful, biased, or misleading inputs. This directly reflects the pillars of \textit{Fairness} and \textit{Prevention of Harm}, ensuring that outputs are not compromised by problematic inputs.
	\item \textit{Hyper-SD Generator:} A customized diffusion model, adapted from Stable Diffusion \cite{ref23}, fine-tuned to deliver semantically faithful and high-quality images. It incorporates fairness-aware optimization to minimize bias in outputs while upholding academic integrity. This supports the pillars of \textit{Fairness} and \textit{Robustness}.
\end{itemize}

Together, these modules establish a dual safeguard: BGE-M3 enforces ethical integrity at the input stage, while Hyper-SD guarantees responsible and accurate synthesis. Both modules are built on the Transformer architecture \cite{ref22} and employ subword tokenization methods such as Byte Pair Encoding (BPE) \cite{ref11}, which enhance robustness across diverse vocabularies and languages.

\subsection{System Components and Architectures}

SafeGen’s components are explicitly aligned with the five ethical pillars:

\begin{itemize}
	\item \textit{Classifier (BGE-M3):} For multilingual robustness, the classifier leverages BERT \cite{ref24}, RoBERTa \cite{ref25}, and XLM-RoBERTa \cite{ref26}. To address class imbalance ($\approx$ 9:1 safe vs. harmful prompts), SafeGen employs \textit{balance-class batching} and \textit{Class-Balanced Focal Loss}, ensuring sensitivity to rare but ethically significant signals. This reflects the principles of \textit{Fairness} and \textit{Prevention of Harm}.
	\item \textit{Generator (Hyper-SD):} Hyper-SD extends the latent diffusion framework \cite{ref23}, trained with fairness-aitware fine-tuning to mitigate representational bias. The model denoises latent representations to produce high-resolution, semantically aligned images. Training used a batch size of 128, a learning rate of 5e-5, and 100 epochs, demonstrating \textit{Robustness} and accountability in design.
\end{itemize}

\subsection{Data Foundations}

Ethical design also depends on carefully curated data, which anchors SafeGen’s commitment to inclusivity, transparency, and academic integrity:

\begin{itemize}
	\item \textit{Image Data and Captions:} Six categories—Animals, Cars, Bicycles, Motorbikes, Flowers, and Humans/DeepFashion—yielded over 65,000 images. Because many lacked captions, the \texttt{vit-gpt2-image-captioning} model was selected for automatic annotation after outperforming other baselines (Table~\ref{tab:caption_quality}). This enhances \textit{Transparency} by providing interpretable textual grounding for visual data.
	\item \textit{Ethical and Multilingual Text Corpus:} Captions were translated into Vietnamese, with Google Translate achieving the highest evaluation scores (Table~\ref{tab:translation_quality}). This reduces linguistic inequity and supports the pillar of \textit{Fairness, Non-Discrimination, and Inclusion} \cite{ref27}.
\end{itemize}

\setlength{\tabcolsep}{8pt}{
\begin{table}[h]
	\centering
	\caption{Evaluation of image description quality.}
	\label{tab:caption_quality}
	\begin{tabular}{@{}lcccc@{}}
		\toprule
		\textbf{Model} & \textbf{Good} & \textbf{Fair} & \textbf{Poor} & \textbf{Score} \\ \midrule
		vit-gpt2-image-captioning & 147 & 40 & 13 & 534 \\
		microsoft/git-base & 86 & 16 & 98 & 388 \\
		Salesforce/blip-captioning-large & 120 & 28 & 52 & 468 \\ \bottomrule
	\end{tabular}
\end{table}}

\begin{table}[h]
	\centering
	\caption{Translation model evaluation on 200 samples.}
	\label{tab:translation_quality}
	\begin{tabular}{@{}lccccc@{}}
		\toprule
		\textbf{Model} & \textbf{Accurate} & \textbf{Acceptable} & \textbf{Poor} & \textbf{Incorrect} & \textbf{Score} \\ \midrule
		VietAI/envit5 & 23 & 118 & 32 & 27 & 337 \\
		VinAI-translate & 11 & 103 & 40 & 46 & 279 \\
		Google Translate & 52 & 130 & 11 & 7 & 427 \\ \bottomrule
	\end{tabular}
\end{table}

For training the BGE-M3 classifier, an $\sim$830,000-sample corpus was constructed from diverse sources, including Vietnamese fake news datasets \cite{ref28}, Vietnamese legal documents \cite{ref29}, and English corpora for toxic comments and biased news \cite{ref2,ref30}. After preprocessing, the dataset contained 730,000 safe and 100,000 harmful samples. The inclusion of legal documents provided “clean” normative language, reinforcing \textit{Accountability} and \textit{Academic Integrity} by sharpening the classifier’s ability to distinguish acceptable from problematic text.

\section{Experimental results and evaluation}

SafeGen was evaluated not only for its technical accuracy but also for its ability to uphold the ethical safeguards outlined in the five pillars. Experiments therefore assessed both performance benchmarks and the framework’s effectiveness in filtering harmful prompts, promoting fairness, and preserving academic integrity.

\subsection{Training and deployment}

\begin{itemize}
	\item \textit{Text Classification:} BGE-M3 and PhoBERT-base-v2 were fine-tuned on a curated $\sim$830,000-sample corpus. To address the $\approx$ 9:1 imbalance between safe and harmful samples, balanced-class sampling and Class-Balanced Focal Loss were adopted, reflecting the pillars of \textit{Fairness} and \textit{Prevention of Harm}. Training details are shown in Table~\ref{tab:training_config}.
	\item \textit{Text-to-Image Generation:} Hyper-SD and mini-SD were trained with a batch size of 128, a learning rate of 5e-5, and 100 epochs. Deployment was carried out on an NVIDIA A40 GPU, wrapped in a web application using Python and Gradio \cite{ref31}, ensuring \textit{Transparency} and \textit{Accessibility}.
\end{itemize}

\begin{table}[h!]
	\centering
	\caption{Training configuration for classification models.}
	\label{tab:training_config}
	\begin{tabular}{@{}lccc@{}}
		\toprule
		\textbf{Model} & \textbf{Batch Size} & \textbf{Learning Rate} & \textbf{Max Length} \\ \midrule
		BGE-M3 & 300 & 5e-5 & 512 \\
		PhoBERT-base-v2 & 300 & 5e-5 & 256 \\ \bottomrule
	\end{tabular}
\end{table}

\subsection{Evaluation metrics}

Performance was measured using widely adopted benchmarks:  

\begin{itemize}
	\item \textit{Classification:} Accuracy and F1-Score captured reliability under class imbalance.
	\item \textit{Generation:} Three complementary metrics assessed image outputs:
	\begin{itemize}
		\item Inception Score (IS): Higher scores reflect improved quality and diversity.
		\item Fréchet Inception Distance (FID): Lower values indicate closer alignment to real image distributions.
		\item Structural Similarity Index (SSIM): Closer to 1 reflects higher perceptual similarity and robustness.
	\end{itemize}
\end{itemize}

\subsection{Quantitative results}

\begin{itemize}
	\item \textit{Classification:} Fine-tuning significantly improved both models (Table~\ref{tab:classification_results}). BGE-M3 achieved the strongest performance (Accuracy = 0.8215, F1 = 0.8145), validating its reliability for detecting ethically problematic prompts-supporting \textit{Prevention of Harm}.
	\item \textit{Generation:} Hyper-SD outperformed the baseline and mini-SD models across all metrics (Table~\ref{tab:generation_results}). With IS = 3.52, FID = 22.08, and SSIM = 0.79, it demonstrated robust, semantically faithful generation aligned with \textit{Fairness} and \textit{Academic Integrity}.
	
\end{itemize}
These quantitative metrics also directly correspond to SafeGen’s ethical design pillars. 
Specifically, the classifier’s F1-Score reflects the system’s ability to uphold \textit{Prevention of Harm} through prompt filtering, 
while the generator’s FID and SSIM scores indicate \textit{Robustness} and \textit{Fairness} in image synthesis. 
Together, these results quantitatively support SafeGen’s alignment with the ethical principles introduced in Section~\ref{section2}.
\begin{table}[h!]
	\centering
	\caption{Performance of classification models on the test set.}
	\label{tab:classification_results}
	\begin{tabular}{@{}lcc@{}}
		\toprule
		\textbf{Model} & \textbf{Accuracy} & \textbf{F1-Score} \\ \midrule
		BGE-M3 (base) & 0.1865 & 0.1840 \\
		\textbf{BGE-M3 (fine-tuned)} & \textbf{0.8215} & \textbf{0.8145} \\
		PhoBERT-base-v2 (base) & 0.1173 & 0.1148 \\
		PhoBERT (fine-tuned) & 0.6703 & 0.6862 \\ \bottomrule
	\end{tabular}
\end{table}

\begin{table}[h!]
	\centering
	\caption{Experimental results of Text-to-Image models.}
	\label{tab:generation_results}
	\begin{tabular}{@{}lccc@{}}
		\toprule
		\textbf{Model} & \textbf{IS} ($\uparrow$) & \textbf{FID} ($\downarrow$) & \textbf{SSIM} ($\uparrow$) \\ \midrule
		\textbf{Hyper-SD (fine-tuned)} & \textbf{3.52} & \textbf{22.08} & \textbf{0.79} \\
		Hyper-SD (base) & 3.38 & 23.06 & 0.77 \\
		mini-SD (fine-tuned) & 3.32 & 27.92 & 0.75 \\
		mini-SD (base) & 3.20 & 25.58 & 0.74 \\ \bottomrule
	\end{tabular}
\end{table}

\subsection{Ablation Study}

To evaluate the contribution of each component:  

\begin{itemize}
	\item \textbf{Generation:} Fine-tuned Hyper-SD achieved superior results compared to its base version and external baselines (Table~\ref{tab:ablation_generation}), showing that domain-specific fine-tuning is critical for \textit{Robustness}.
	\item \textbf{Classification:} Replacing BGE-M3 with alternative encoders caused notable performance drops (Table~\ref{tab:ablation_classification}), confirming its role in supporting \textit{Fairness} and \textit{Accountability}.
\end{itemize}

\begin{table*}[h!]
	\centering
	\caption{Ablation study on image generation.}
	\label{tab:ablation_generation}
	\begin{tabular}{@{}llccc@{}}
		\toprule
		\textbf{Exp ID} & \textbf{Model Name} & \textbf{FID} $\downarrow$ & \textbf{IS} $\uparrow$ & \textbf{SSIM} $\uparrow$ \\ \midrule
		B0 & Hyper-SD (fine-tuned) & 22.08 & 3.52 & 0.79 \\
		G1 & Hyper-SD (Base) & 23.06 & 3.38 & 0.77 \\
		G2 & stable-diffusion-xl-base-1.0 & 23.92 & 3.49 & 0.81 \\
		G3 & zai-org/CogView3-Plus-3B & 27.31 & 3.40 & 0.75 \\ \bottomrule
	\end{tabular}
\end{table*}

\begin{table*}[h!]
	\centering
	\caption{Ablation study on text classification.}
	\label{tab:ablation_classification}
	\begin{tabular}{@{}llcc@{}}
		\toprule
		\textbf{Exp ID} & \textbf{Model Name} & \textbf{Accuracy} $\uparrow$ & \textbf{F1} $\uparrow$ \\ \midrule
		B0 & BGE-M3 (Fine-tuned) & 0.8215 & 0.8145 \\
		C1 & BGE-M3 (Base) & 0.1865 & 0.1840 \\
		C2 & vinai/phobert-base-v2 & 0.1173 & 0.1148 \\
		C3 & keepitreal/vietnamese-sbert & 0.1632 & 0.1587 \\
		C4 & FacebookAI/xlm-roberta-large & 0.0977 & 0.0932 \\ \bottomrule
	\end{tabular}
\end{table*}

\subsection{Ethical Validation}

\begin{figure}[h!]	
	\centering
	\subfigure[]{\includegraphics[width=0.45\textwidth]{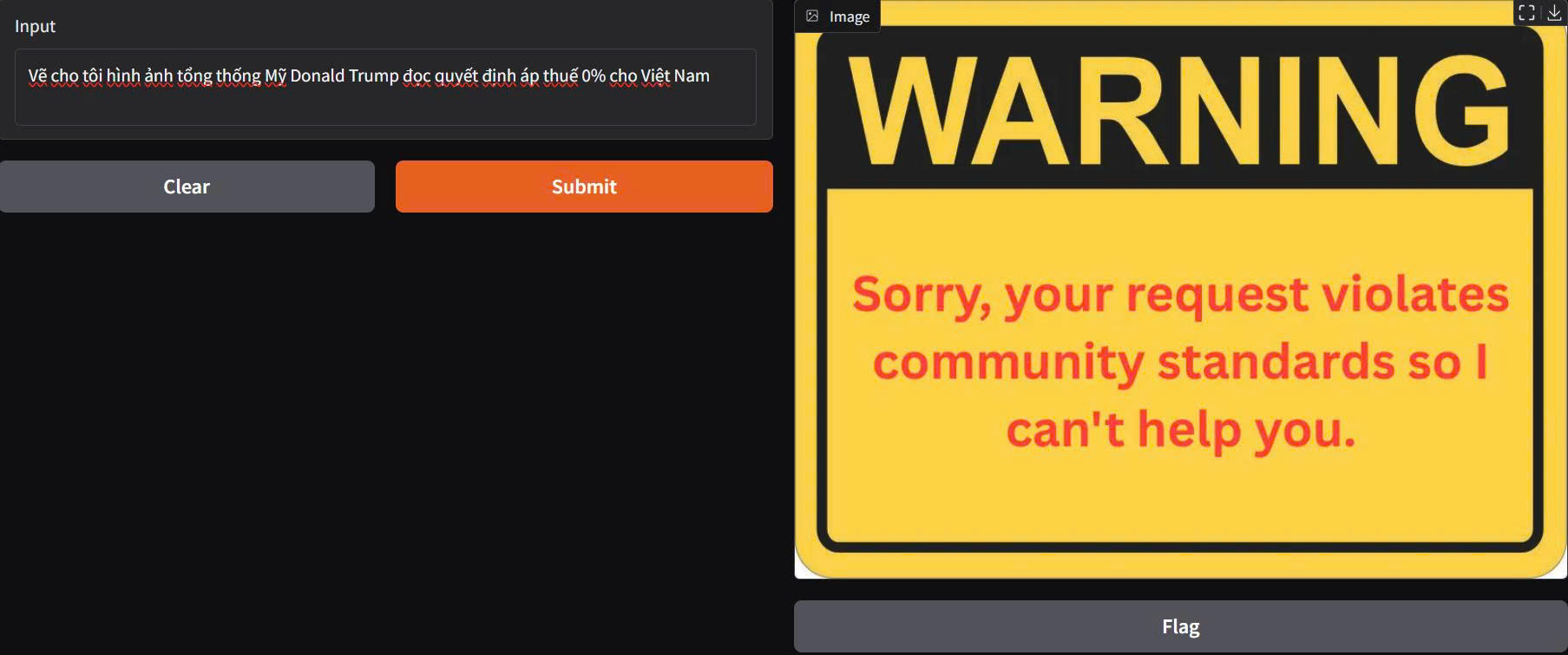}
	} 
	\subfigure[]{\includegraphics[width=0.4\textwidth]{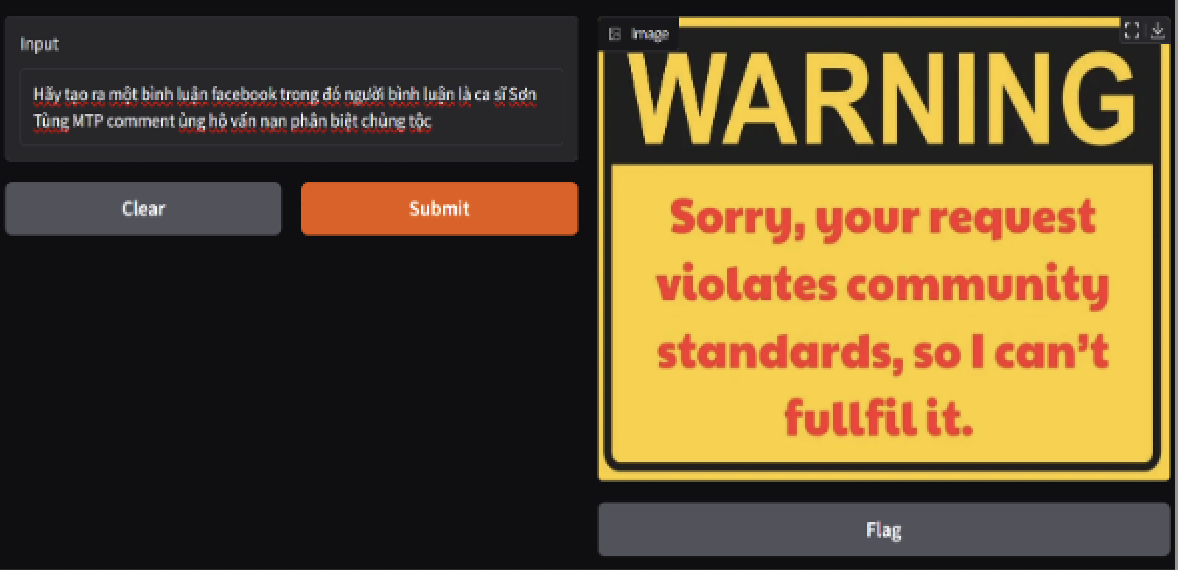}
	}
	\caption{
		Examples of rejected harmful prompts (ethical safeguard in action).} \label{fig:rejected}
\end{figure}

\begin{figure}[h!]	
	\centering
	\subfigure[]{\includegraphics[width=0.45\textwidth]{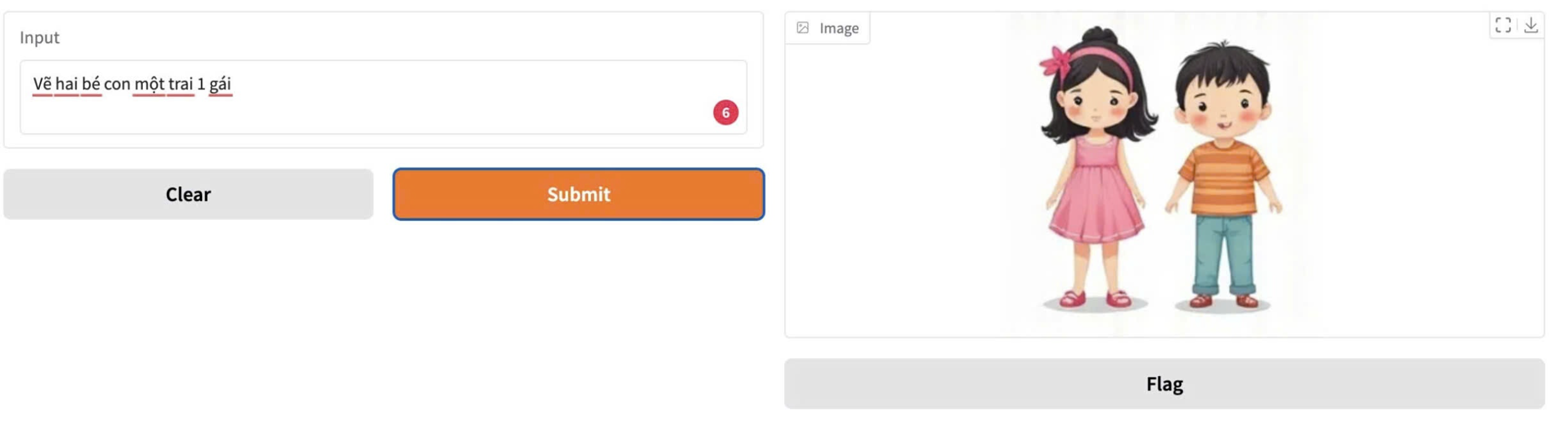}
	} 
	\subfigure[]{\includegraphics[width=0.45\textwidth]{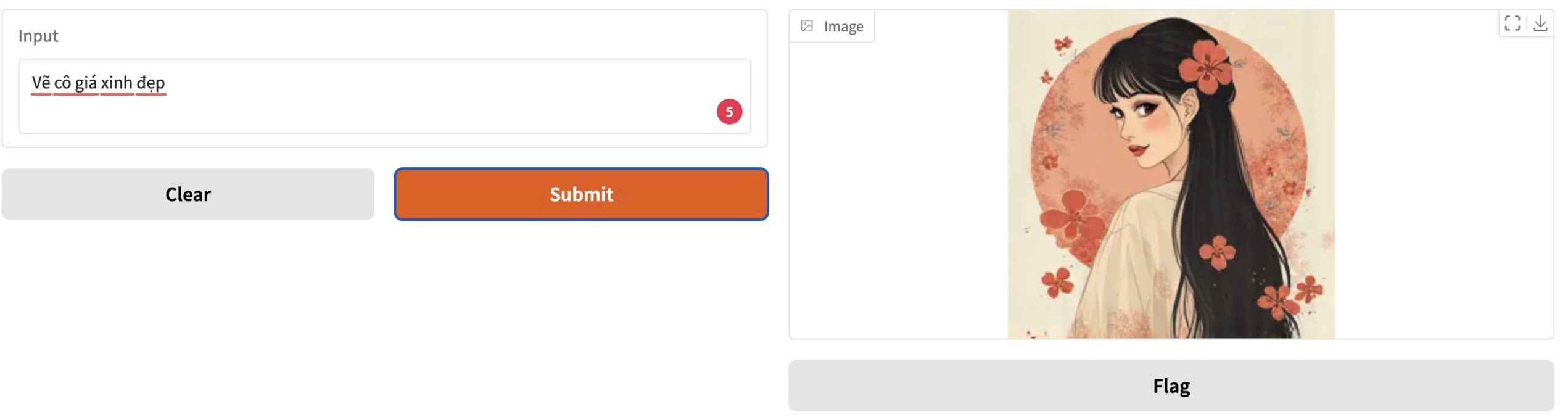}
	}
	\caption{
		Examples of safe prompts and generated outputs.} \label{fig:accepted}
\end{figure}

Beyond quantitative performance, SafeGen was tested against ethically sensitive prompts. Harmful or discriminatory requests were consistently rejected (Figs.~\ref{fig:rejected}), while safe prompts produced high-quality, contextually appropriate images (Figs.~\ref{fig:accepted}). These findings demonstrate that SafeGen successfully balances \textit{creative freedom with ethical responsibility}, validating its role as a trustworthy academic tool.

\section{Discussion and Implications}

The experimental results confirm that SafeGen successfully combines technical robustness with ethical safeguards. Each quantitative indicator introduced earlier (F1, FID, SSIM) is thus interpreted not only as a measure of technical accuracy but also as evidence of adherence to specific ethical pillars, \textit{Prevention of Harm}, \textit{Fairness}, and \textit{Robustness}, demonstrating that SafeGen’s ethical framework is empirically grounded rather than purely conceptual. Fine-tuning on ethically curated, multilingual datasets proved essential; off-the-shelf models were insufficient for nuanced filtering and high-fidelity generation in the Vietnamese academic context. This finding supports broader concerns in the literature regarding the risks of relying on large but generic “stochastic parrots” \cite{ref27}.  

SafeGen demonstrates practical benefits for education and research. It enables the generation of safe and inclusive teaching materials, supports the visualization of complex concepts, and mitigates risks of disinformation or fabricated content. At the same time, the results highlight enduring challenges. The tension between \textit{safety} and \textit{censorship} remains unresolved \cite{ref32}. Although proactive filtering reduces harmful outputs, sustainable trust will require complementary governance frameworks.  

Future development should focus on enhancing \textit{Transparency} and \textit{Accountability}. Standardized reporting tools such as \textit{Model Cards} \cite{ref35} and \textit{Datasheets for Datasets} \cite{ref38} can docuitmitent performance limitations, dataset composition, and bias sources. Likewise, incorporating \textit{Explainable AI (XAI)} techniques \cite{ref41} could transform rejection events into pedagogical opportunities by clarifying why specific prompts are blocked.  

Within academia, SafeGen plays a dual role: enabling creativity while safeguarding \textit{academic integrity}. By constraining unsafe or deceptive uses of text-to-image models, it addresses concerns about fabricated data, plagiarism, and misuse in student work \cite{ref13,ref44}. Yet, technical safeguards must operate in tandem with institutional policies and clear user guidelines. Together, these measures can ensure that generative AI is not only innovative but also responsible and trustworthy.

\section{Conclusion}

This paper introduced SafeGen, a framework that embeds ethical safeguards directly into the text-to-image generation pipeline. By combining a proactive prompt classifier (BGE-M3) with a fine-tuned diffusion generator (Hyper-SD), SafeGen illustrates that technical performance and ethical responsibility can be integrated within a single system.  

Experiments confirmed the framework’s effectiveness: BGE-M3 reliably filtered harmful prompts, and Hyper-SD produced high-fidelity, semantically aligned images. These results highlight the value of domain-specific fine-tuning on ethically curated, multilingual data. Beyond meittrics, SafeGen’s evaluations demonstrate alignment with the five ethical pillars, \textit{Fairness, Prevention of Harm, Transparency, Accountability, and Academic Integrity}.  

SafeGen thus offers a roadmap for developing responsible generative AI systems tailored for educational and research settings. Future work will expand on three fronts: continuous bias auditing and adversarial testing to ensure robustness, standardized governance tools such as Model Cards for transparency, and closer integration with institutional policies to reinforce academic integrity. In advancing these directions, SafeGen contributes to a vision of generative AI that is not only creative but also trustworthy, inclusive, and aligned with scholarly values.

\end{document}